\documentclass{article}

\usepackage[preprint]{neurips_2025}

\usepackage{graphicx}

\usepackage{float}
\usepackage{longtable}

\usepackage{amsmath}

\usepackage[utf8]{inputenc} 
\usepackage[T1]{fontenc}    
\usepackage{hyperref}       
\usepackage{url}            
\usepackage{booktabs}       
\usepackage{amsfonts}       
\usepackage{nicefrac}       
\usepackage{microtype}      
\usepackage{xcolor}         

\title{Unknown Aware AI-Generated Content Attribution}

\author{
  Ellie Thieu \\
  UW--Madison \\
  \texttt{thieu@wisc.edu}
  \And
  Jifan Zhang \\
  UW--Madison \\
  \texttt{jifan@cs.wisc.edu}
  \And
  Haoyue Bai \\
  UW--Madison \\
  \texttt{haoyue.bai@wisc.edu}
}

\begin{document}

\maketitle

\begin{abstract}

The rapid advancement of photorealistic generative models has made it increasingly important to attribute the origin of synthetic content, moving beyond binary real or fake detection toward identifying the specific model that produced a given image. We study the problem of distinguishing outputs from a target generative model (e.g., OpenAI’s DALL·E 3) from other sources, including real images and images generated by a wide range of alternative models.
Using CLIP features and a simple linear classifier, shown to be effective in prior work, we establish a strong baseline for target generator attribution using only limited labeled data from the target model and a small number of known generators. However, this baseline struggles to generalize to harder, unseen, and newly released generators. To address this limitation, we propose a constrained optimization approach that leverages unlabeled wild data, consisting of images collected from the Internet that may include real images, outputs from unknown generators, or even samples from the target model itself. The proposed method encourages wild samples to be classified as non target while explicitly constraining performance on labeled data to remain high.
Experimental results show that incorporating wild data substantially improves attribution performance on challenging unseen generators, demonstrating that unlabeled data from the wild can be effectively exploited to enhance AI generated content attribution in open world settings.

\end{abstract}

\section{Introduction}

Recent advances in deep generative modeling have led to dramatic improvements in
image synthesis quality~\citep{song2019generative, ramesh2021zero,
rombach2022high, vahdat2021score, ramesh2022hierarchical}. While these advances
have expanded the range of applications for generative models, they have also
raised significant concerns about misuse. AI-generated content
detection, aimed at distinguishing synthetic images from real ones, has become a
critical task and has attracted substantial research attention, leading to a
variety of proposed methods~\cite{chai2020makes, wang2020cnn,
ojha2023towards, bammey2024synthbuster}. Beyond detection, however, attributing
the origin of synthetic content is equally important, as it identifies which
specific generative model produced a given image. Such attribution enables
provenance tracking and supports accountability by linking content back to the
responsible model developer. For example, determining whether an image was
generated by DALL·E~3~\citep{betker2023improving}, rather than by Midjourney or
Stable Diffusion~\citep{rombach2022high}, has direct implications for governance,
transparency, and safety in generative AI.

The rapid and continuous release of new generative models poses a fundamental
challenge to existing attribution methods. This fast-paced evolution makes it
impractical to rely on supervised pipelines that assume all possible generators
are known at training time. As a result, a critical and largely unsolved
question arises:

\begin{center}
    \textbf{\emph{Can we design a learning framework for target generator
    attribution that remains robust in the presence of unknown or newly
    released generators?}}
\end{center}

A common approach is to train a classifier using labeled examples from the target
generator along with a small set of known non-target sources. Prior work has
shown that CLIP features~\citep{radford2021learning}, combined with a simple
linear classifier, can achieve strong attribution performance using only limited
labeled data from the target generator and a few known generators~\citep{
ojha2023towards}. However, such models often struggle to generalize to more
challenging, newly released generators. Moreover, collecting labeled data for
every possible non-target generator is infeasible in practice, given the
constant emergence of new models. In contrast, unlabeled images from the wild,
such as those collected from the Internet, are abundant, diverse, and
inexpensive to obtain.

\begin{figure}
\centering
\includegraphics[width=0.65\textwidth]{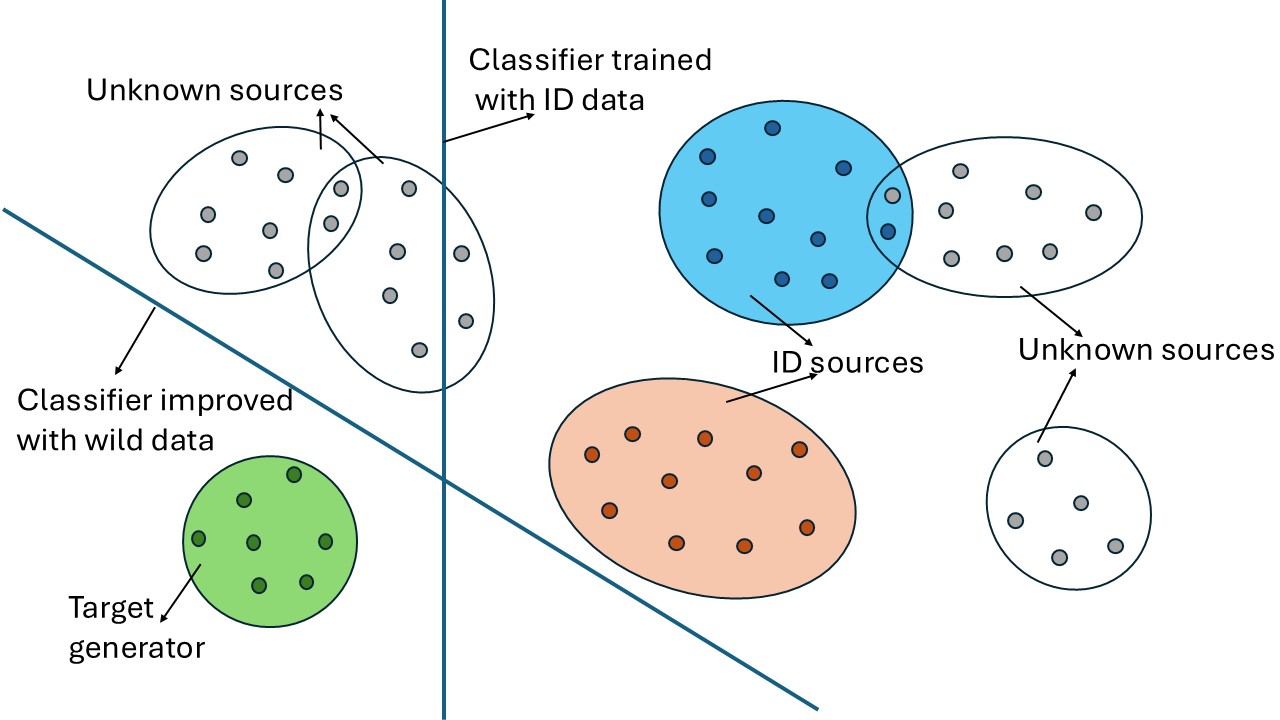}
\caption{Illustration of unknown-aware generator attribution.
A classifier trained only on labeled sources learns a
decision boundary that separates the target generator from known sources but may
misclassify samples from unknown or unseen generators. By incorporating
unlabeled wild data under a constrained optimization framework, the decision
boundary is adjusted to better separate the target generator from diverse
unknown sources, while preserving performance on labeled source data.}
\end{figure}
\label{fig:teaser}

To address this challenge, we propose an unknown-aware AI-generated content
attribution framework that fine-tunes a given classifier using unlabeled wild
data through a constrained optimization procedure. The wild data may include
real images, outputs from unknown generators, or even content produced by the
target generator itself, and is treated as non-target to encourage
generalization. A constraint is enforced to preserve performance on the
original labeled data, allowing the model to benefit from the diversity of wild
data while maintaining accuracy on known sources. Experimental results show
that our method consistently improves attribution of DALL·E~3 images,
particularly against challenging and unseen generators such as Midjourney,
Firefly, and Stable Diffusion~XL. These results demonstrate that unlabeled wild
data can be effectively leveraged to build scalable and robust AI content
attribution systems in open-world settings.

Figure~\ref{fig:teaser} illustrates the key intuition behind the proposed
framework. The vertical line represents the decision boundary learned by a
baseline classifier trained only on ID data, while the adjusted boundary shows
the effect of fine-tuning with unlabeled wild data under the proposed
constraint.

Our main contributions are summarized as follows:
\begin{itemize}
\item \textbf{A constrained fine-tuning framework for unknown-aware attribution.}
We introduce a simple and stable constrained optimization approach that
fine-tunes a classifier using unlabeled wild data while explicitly preserving
performance on labeled in-distribution data, preventing collapse and
catastrophic forgetting.

\item \textbf{Leveraging unlabeled wild data for improved generalization.}
We propose a training framework that incorporates unlabeled Internet images,
despite their noise and diversity, by enforcing a constraint that maintains
performance on trusted labeled data while improving robustness to unseen
generators.

\item \textbf{Problem formulation for target generator attribution.}
We study the fine-grained binary task of determining whether an image was
generated by a specific model (e.g., DALL·E~3), moving beyond real-or-fake
detection to support accountability and AI governance. Given current industry
practices, single-target attribution is more practical and societally relevant
than multi-class attribution across many generators.

\item \textbf{Unknown-aware attribution in open-world settings.}
We highlight and empirically study the challenge of identifying target model
outputs in the presence of unknown or newly released generators, a realistic
yet underexplored setting in the AI-generated content literature.
\end{itemize}

\section{Related Work}

\subsection{Synthetic Image Generation}

Generative modeling for images has progressed rapidly over the past decade.
Early advances were driven by Generative Adversarial Networks (GANs), which
enabled high-quality image synthesis across a range of domains
\cite{goodfellow2014generative, choi2018stargan, zhu2017unpaired, brock2018large, bai2023improving}.
Subsequent work explored transformer-based architectures to further improve
image generation fidelity and scalability
\cite{parmar2018image, zhang2022styleswin, chang2023muse}.

More recently, diffusion models have emerged as the dominant paradigm for image
generation, substantially advancing the state of the art and leading to widely
used systems such as Stable Diffusion~\cite{rombach2022high}, DALL·E and
DALL·E~2~\cite{ramesh2022hierarchical}, DALL·E~3~\cite{betker2023improving},
GLIDE~\cite{nichol2021glide}, and related variants
\cite{dhariwal2021diffusion, zhu2023conditional}. This ecosystem continues to
evolve rapidly, with new generators frequently introduced and existing models
iteratively refined. As a result, downstream detection and attribution systems
must contend with a continuously shifting landscape of generative techniques,
many of which may be unavailable or unknown at training time.

\subsection{AI-Generated Content Detection and Attribution}

The rapid improvement of generative models has motivated extensive research on
AI-generated content detection and attribution. Early detection methods focused
primarily on GAN-generated images and relied on handcrafted visual cues. For
example, prior work examined facial artifacts such as eyes and teeth
\cite{matern2019exploiting} or low-level color inconsistencies
\cite{mccloskey2018detecting}. Subsequent approaches sought generation-specific
signatures, including frequency-domain artifacts
\cite{marra2019gans, zhang2019detecting}, GAN fingerprints
\cite{yu2019attributing}, and localized artifacts captured through limited
receptive fields~\cite{chai2020makes}.

With the rise of diffusion-based generators, many GAN-oriented detectors were
shown to degrade substantially~\cite{corvi2023detection}. In response, recent
methods have proposed diffusion-specific detection strategies~\cite{cozzolino2018forensictransfer, bai2025s}. For instance,
Synthbuster~\cite{bammey2024synthbuster} exploits Fourier statistics of residuals,
DE-FAKE~\cite{sha2023fake} incorporates prompt information, and
DIRE~\cite{wang2023dire} leverages reconstruction behavior in diffusion models.
Other approaches analyze physical inconsistencies such as lighting or
perspective~\cite{farid2022lighting, farid2022perspective}. Universal detectors
based on large vision transformers and CLIP representations have also been
proposed~\cite{ojha2023towards}, though these methods are typically trained using
labeled data from known generators.

Despite these advances, most existing approaches implicitly assume a closed-world
setting in which the set of generators encountered at test time is known during
training. As demonstrated in prior work~\cite{cozzolino2018forensictransfer},
detectors trained on specific architectures or generators often fail to
generalize across model families, highlighting the limitations of relying on
model-specific artifacts.

In contrast, we study \emph{unknown-aware} AI-generated content attribution, a
more realistic setting in which test-time content may originate from unseen or
newly released generators. Rather than assuming access to labeled data from all
possible sources, our approach leverages unlabeled, in-the-wild image mixtures to
improve robustness to evolving generative models.

\subsection{Open-Set Attribution}

A small body of prior work has considered open-set settings for image
attribution, though most focus on GANs or operate at the architectural level.
For example, Abady et al.~\cite{abady2024siamese} address open-set attribution by
identifying generator architectures rather than specific models. Other works
that perform generator-level attribution in open-world settings
\cite{girish2021openworld, wang2023vit} primarily target GAN-based generators,
which are generally less challenging than modern diffusion models or proprietary
commercial systems.

To handle unseen generators, some approaches introduce an explicit rejection or
``unknown'' class~\cite{yang2023progressive, wang2025bosc}. While effective at
avoiding overconfident misclassification, these methods emphasize correct
classification among known generators rather than improving generalization to
challenging unseen ones. In contrast, our approach focuses on leveraging unlabeled
wild data to directly improve attribution robustness to unknown generators,
without requiring explicit rejection modeling or labeled examples from unseen
sources.

\subsection{Semi-Supervised Learning}

Semi-supervised learning (Semi-SL) provides a general framework for leveraging
unlabeled data alongside labeled examples in classification tasks
\cite{zhu2005semi, vanEngelen2020, ouali2020overview}. Most modern Semi-SL methods
are built upon two core principles: \emph{consistency regularization} and
\emph{pseudo labeling}~\cite{Sohn2020FixMatchSS, Berthelot2020ReMixMatch}.
Consistency-based approaches encourage prediction invariance under data
augmentation, while pseudo-labeling methods assign labels to unlabeled examples
with high-confidence predictions. Extensions such as FlexMatch
\cite{Zhang2021FlexMatchBS}, FreeMatch~\cite{wang2022freematch}, and SoftMatch
\cite{chen2023softmatch} further refine these ideas through adaptive thresholding
and confidence calibration.

Although Semi-SL methods have achieved strong results on natural image
classification benchmarks, their assumptions often break down in the context of
AI-generated image detection and attribution. In particular, strong data
augmentations may suppress or distort subtle generation artifacts that are
critical for identifying the source model. As a result, enforcing consistency
across augmented views can be detrimental in this setting.

Recent work has explored Semi-SL with large pretrained models and vision
transformers~\cite{cai2022semisupervised, lagunas2023transfer, xing2023svformer},
and benchmarks such as USB~\cite{wang2022usb} provide standardized evaluations.
However, these approaches do not explicitly address the challenges posed by synthetic image attribution under evolving generative distributions. Our work is also closely related to semi-supervised novelty detection \cite{blanchard2010semi} and its extensions to out-of-distribution (OOD) detection \cite{katz2022training, bai2023feed, bai2024aha, bai2024out}. These methods aim to identify novel samples from mixtures of known and unknown distributions. We draw inspiration from this line of work, adapting its principles to the problem of unknown-aware generator attribution.

\section{Problem Setup}

We study the problem of \emph{targeted generator attribution}: given an image
$\mathbf{x}$, determine whether it was produced by a specific target generative
model $\mathbb{G}_t$ (e.g., DALL·E~3). This task can be formulated as a binary
classification problem, where the goal is to decide whether
$\mathbf{x} \sim \mathbb{P}_{\mathbb{G}_t}$ or whether it originates from any
non-target source, including other generative models
$\mathbb{G} \neq \mathbb{G}_t$ or real images.

\paragraph{Labeled in-distribution data.}
Let $\mathcal{X}=\mathbb{R}^d$ denote the input space and
$\mathcal{Y}=\{0,1\}$ the label space, where $y=1$ indicates that an image was
generated by the target model $\mathbb{G}_t$, and $y=0$ denotes non-target
content. We assume access to a labeled in-distribution (ID) training dataset
$\mathcal{D}_{\text{labeled}}
= \{(\mathbf{x}_i, y_i)\}_{i=1}^n$,
where each sample $\mathbf{x}_i$ is drawn either from the target generator
distribution $\mathbb{P}^{\mathcal{X}}_{\mathbb{G}_t}$ with label $y_i=1$, or
from a set of known non-target sources with label $y_i=0$. These known sources
may include real images and images generated by a limited number of auxiliary
generative models available at training time.

\paragraph{Unlabeled wild data.}
At deployment time, the classifier encounters inputs drawn from a broader and
uncontrolled mixture of sources. We model this setting via a \emph{wild}
distribution,
\[
\mathbb{P}_{\text{wild}}
:= (1-\pi_k-\pi_u)\,\mathbb{P}^{\mathcal{X}}_{\mathbb{G}_t}
\;+\;
\pi_k\,\mathbb{P}_{\text{known}}
\;+\;
\pi_u\,\mathbb{P}_{\text{unknown}},
\]
where $\pi_k,\pi_u \ge 0$ and $\pi_k+\pi_u \le 1$ are unknown mixture proportions.
Here:
\begin{itemize}
    \item \textbf{Target distribution} $\mathbb{P}^{\mathcal{X}}_{\mathbb{G}_t}$
    denotes the marginal distribution of images generated by the target model
    $\mathbb{G}_t$;
    \item \textbf{Known-source distribution} $\mathbb{P}_{\text{known}}$
    corresponds to non-target sources observed during training;
    \item \textbf{Unknown-source distribution} $\mathbb{P}_{\text{unknown}}$
    represents images from previously unseen generators or other novel sources.
\end{itemize}

This formulation captures a realistic open-world deployment scenario in which
test-time inputs may originate from both seen and unseen generative processes.
Depending on the training setup, real images may appear in either the known or
unknown source distributions. The central challenge is to learn a decision
function that reliably separates target-model outputs from all other sources,
despite uncertainty about the composition of the wild data.

\paragraph{Learning framework.}
Let $f_\theta : \mathcal{X} \rightarrow \mathbb{R}$ denote a scoring function
parameterized by $\theta$, where higher scores indicate a higher likelihood that
an input is generated by the target model. We first train $f_\theta$ using the
labeled ID dataset $\mathcal{D}_{\text{labeled}}$. Subsequently, the model may be
fine-tuned using unlabeled wild data drawn from $\mathbb{P}_{\text{wild}}$ to
improve robustness to unknown generators. Performance is evaluated using threshold-independent metrics, including Average Precision (AP) and the Area Under the ROC Curve (AUROC).

\section{Unknown-Aware AI-Generated Content Attribution}
\label{sec:method}

We introduce an unknown-aware framework for generator-specific image attribution
that remains robust in the presence of unseen or newly released generators. The
task is to determine whether a given image was generated by a specific target
model. In our experiments, we focus on OpenAI’s DALL·E~3 as a representative case
study. While we use DALL·E~3 throughout for concreteness, the proposed framework
is model-agnostic and can be readily applied to any target generator.

Our approach follows a two-stage procedure. We first train a baseline attribution
classifier using a small labeled dataset consisting of images from the target
generator and a limited set of known non-target sources. While this baseline
achieves strong performance on in-distribution (ID) data, it often fails to
generalize to images produced by unseen or newly released generators. To address
this limitation, we introduce a constrained fine-tuning strategy that leverages
unlabeled wild data. The key idea is to expose the classifier to a broad and
diverse distribution of images while explicitly constraining performance on
trusted labeled data, thereby improving generalization without sacrificing
in-distribution accuracy.

We adopt a binary attribution formulation tailored to practical deployment
scenarios in which an organization seeks to determine whether content was
generated by its own proprietary model. Although our experiments focus on this
single-target setting, the proposed constrained optimization framework is
general and can be extended to multi-generator attribution or joint detection
and attribution tasks with minimal modification.

In the remainder of this section, we first describe the baseline classifier
trained on labeled ID data (Section~\ref{Training baseline classifier}), and then
present our constrained fine-tuning approach that incorporates unlabeled wild
data (Section~\ref{constrained optimization}).

\subsection{Training on In-Distribution Data}
\label{Training baseline classifier}

We begin by training a baseline attribution classifier using labeled
in-distribution (ID) data consisting of images from the target generator
(DALL·E~3) and a small set of available non-target generators. We denote this
dataset by $D_{\text{labeled}}$. Each image is encoded using the CLIP ViT-L/14
image encoder, from which we extract a 768-dimensional feature vector from the
penultimate layer. CLIP features have been shown to be highly effective for
synthetic image detection and attribution tasks~\cite{ojha2023towards}, and
recent work suggests that similar performance can be obtained with alternative
CLIP variants~\cite{Cozzolino2024raisethebar}.

On top of the frozen CLIP features, we train a lightweight linear classifier for
binary attribution. Images generated by the target model are assigned label $0$,
and all other images are assigned label $1$. The classifier consists of a single
linear layer followed by a sigmoid activation, producing the probability that an
image does \emph{not} originate from the target generator. Training is performed
using binary cross-entropy (BCE) loss and the Adam optimizer, with early stopping
based on validation loss computed on a held-out subset of $D_{\text{labeled}}$.

The resulting classifier serves as our baseline model. We record its BCE loss on
the labeled ID data, which later provides a reference point for constraining
performance during fine-tuning with wild data.

\subsection{Fine-Tuning with Wild Data via Constrained Optimization}
\label{constrained optimization}

To improve generalization to unseen and challenging generators, we incorporate
unlabeled wild data collected from diverse sources such as the Internet. This
wild data may include real images, images generated by known non-target models,
images from previously unseen generators, and potentially a small fraction of
images produced by the target generator itself. As ground-truth labels are
unavailable, these samples cannot be used in a fully supervised manner.

Rather than explicitly assigning hard labels to wild samples, we treat wild data
as an auxiliary signal that encourages the classifier to expand its decision
boundary away from the target generator. Naïvely optimizing on wild data alone
would lead to degenerate solutions (e.g., predicting all samples as non-target).
To prevent this, we introduce a constrained optimization formulation that
explicitly preserves performance on labeled ID data while leveraging the
diversity of wild samples.

Concretely, we fine-tune the baseline classifier using both labeled ID data and
unlabeled wild data, while enforcing a constraint on the loss over
$D_{\text{labeled}}$. This constraint ensures that fine-tuning does not degrade
the classifier’s original attribution capability. In all experiments, we set
the constraint threshold to twice the loss achieved by the baseline classifier
trained solely on labeled ID data.

The same CLIP feature extraction pipeline is applied to the wild data. Once
features are extracted, fine-tuning is performed using a constrained objective
described below. Because the classifier is a low-capacity linear model over
fixed representations, this procedure remains stable even when the wild dataset
is large or highly diverse.

\paragraph{Learning Objective}
Our goal is to improve attribution robustness to unknown or unseen generators by
leveraging unlabeled wild data, while explicitly preserving performance on labeled
in-distribution (ID) data.
Let $D_{\text{labeled}}=\{(x_j,y_j)\}_{j=1}^n$ denote the labeled ID dataset, where
$y_j\in\{0,1\}$ indicates whether $x_j$ is generated by the target model ($0$) or
not ($1$). Let $D_{\text{wild}}=\{\tilde{x}_i\}_{i=1}^m$ denote the unlabeled wild
dataset. Let $f_\theta(x)$ denote the classifier output, and let
$L(\cdot,\cdot)$ denote the binary cross-entropy (BCE) loss.

We formulate the fine-tuning objective as the following constrained optimization
problem:
\begin{align}
\min_{\theta} \;& \frac{1}{m}\sum_{i=1}^m L\!\left(f_\theta(\tilde{x}_i), 1\right) \\
\text{s.t. } \;& \frac{1}{n}\sum_{j=1}^n L\!\left(f_\theta(x_j), y_j\right) \le \alpha ,
\end{align}
where the objective encourages wild samples to be classified as \emph{non-target},
while the constraint ensures that the average loss on labeled ID data does not
exceed a predefined threshold $\alpha$. In all experiments, we set $\alpha$ to
twice the BCE loss achieved by the classifier trained solely on
$D_{\text{labeled}}$.

In practice, we optimize a Lagrangian relaxation of the constrained problem by
minimizing a weighted sum of the two losses:
\begin{align}
\mathcal{L}(\theta)
= \frac{1}{n}\sum_{j=1}^n L\!\left(f_\theta(x_j), y_j\right)
+ \lambda \frac{1}{m}\sum_{i=1}^m L\!\left(f_\theta(\tilde{x}_i), 1\right),
\end{align}
where $\lambda \ge 0$ controls the strength of the wild-data regularization. The
value of $\lambda$ is chosen such that the resulting labeled ID loss remains close
to the constraint threshold $\alpha$, thereby preventing degenerate solutions that
would trivially classify all samples as non-target.

During fine-tuning, we monitor the compound loss on a held-out validation set drawn
from the same distributions as the training data and apply early stopping once the
loss stabilizes. This procedure allows the classifier to benefit from the diversity of wild data while maintaining reliable performance on known in-distribution sources.

\section{Experiments}

Our experimental design reflects a realistic open-world deployment setting. We
train an attribution classifier using images from a small set of currently
available generators, and evaluate its ability to generalize to newer and more
challenging generators that were not observed during training. This setup
captures two practical constraints encountered in real-world attribution
systems: (i) the set of generative models evolves rapidly, and (ii) collecting
large, labeled attribution datasets is costly, whereas unlabeled images from the
wild are abundant.

Throughout our experiments, we focus on OpenAI’s DALL·E~3 as the target generator
and train a classifier to distinguish images generated by DALL·E~3 from all
other sources using CLIP features and a linear classifier. We evaluate
attribution performance on three categories of sources: (1) generators observed
during labeled training (in-distribution), (2) generators that appear only in
the wild data used for fine-tuning, and (3) generators that appear in neither
the labeled training set nor the wild data. Results are reported both with and
without incorporating wild data.

\subsection{Experimental Details}

\paragraph{Datasets.}
For the main experiment, the labeled in-distribution (ID) training set consists
of images from four generators: DALL·E~3, Wukong, Stable Diffusion~v1.4, and
Stable Diffusion~v1.5, with DALL·E~3 serving as the target generator. We use 200
images from DALL·E~3 and 67 images from each auxiliary generator, yielding a
balanced binary classification dataset. This configuration demonstrates that
strong attribution performance can be achieved with limited labeled data when
leveraging pretrained CLIP representations, while reflecting realistic data
collection constraints.

To simulate unlabeled wild data, we construct a dataset containing 67 images
from each of the following sources: ImageNet (real images), DALL·E~3, ADM,
BigGAN, GLIDE (4.5B), Midjourney, Stable Diffusion~v1.4, Stable
Diffusion~v1.5, VQDM, Wukong, Firefly, Stable Diffusion~XL,
LDM\_200\_cfg, and DALL·E~2. The inclusion of DALL·E~3 images in the wild set
introduces label noise, reflecting realistic scenarios in which wild data may
contain unlabeled samples from the target generator.

\paragraph{Data Sources.}
DALL·E~3 images are obtained from the Hugging Face DALL·E~3 dataset.\footnote{\url{https://huggingface.co/datasets/OpenDatasets/dalle-3-dataset}}
Real images are sourced from ImageNet.\footnote{\url{https://www.kaggle.com/c/imagenet-object-localization-challenge/overview/description}}

Images from Midjourney, Stable Diffusion~v1.4 and v1.5, ADM, GLIDE (4.5B), Wukong,
VQDM, and BigGAN are obtained from the GenImage dataset~\citep{zhu2023genimage}.
Additional image sources—including GLIDE\_50\_27, GLIDE\_100\_10,
GLIDE\_100\_27, Guided, LDM\_100, LDM\_200\_cfg, and LDM\_200, are taken from
\citep{ojha2023towards}. From Synthbuster~\citep{bammey2024synthbuster}, we
incorporate DALL·E~2, Firefly, additional Midjourney samples, and Stable
Diffusion variants v1.3, v1.4, v2, and XL.

The full dataset is aggregated from multiple sources, resulting in substantial
variation in the number of available images per generator, ranging from
approximately 1{,}000 to over 160{,}000. While we curate a diverse and
up-to-date collection spanning both open-source and proprietary generators, the
rapid pace of model development implies that any fixed dataset will inevitably
become outdated. This further motivates the need for unknown-aware attribution
methods.

For generators included in the wild set, we randomly sample 67 images per source
for training. Evaluation is performed using 600 test images per generator for
all sources, regardless of whether they appear in the training or wild sets, to
ensure consistency across evaluations.

We report Average Precision (AP) and Area Under the ROC Curve (AUROC), which are
threshold-independent and better reflect attribution performance than accuracy.
AP and AUROC are computed by concatenating classifier outputs for DALL·E~3 test
images and images from a given comparison source, together with their
ground-truth labels, and applying
\texttt{average\_precision\_score} and \texttt{roc\_auc\_score} from
\texttt{scikit-learn}.

\paragraph{Model Architecture.}
We use CLIP ViT-L/14 as the image encoder and extract a 768-dimensional feature
vector from the penultimate layer for each image. On top of these frozen
representations, we train a lightweight linear classifier consisting of a single
fully connected layer followed by a sigmoid activation. The classifier outputs
the probability that an image is \emph{not} generated by DALL·E~3.

The classifier is trained using the Adam optimizer with a learning rate of
$10^{-3}$ and binary cross-entropy loss. We use large batch sizes so that each
training step aggregates gradients over all available labeled and wild samples.
Early stopping is applied based on validation loss. During fine-tuning with wild
data, we optimize the compound loss described in
Section~\ref{sec:method}, adjusting the wild-data loss weight $\lambda$ to
maintain performance on labeled ID data. Training remains stable across runs due
to the low-capacity linear classifier and explicit control of $\lambda$.

The dominant computational cost arises from CLIP feature extraction, which is
GPU-accelerated and completes within a few hours even for large datasets. Once
features are extracted, training the linear classifier is highly efficient and
typically completes within minutes.

\subsection{Main Results}

main experiment. 
Table~\ref{tab:main_results_short} summarizes attribution
performance on in-distribution (ID) generators and a subset of challenging
generators, both before and after incorporating wild data.

Without wild data, the baseline classifier already achieves strong performance
on most sources, often exceeding 99\% in both Average Precision (AP) and ROC AUC.
However, several generators—most notably Midjourney, Firefly, and Stable
Diffusion~XL—remain substantially more challenging, exhibiting noticeably lower
scores. This observation is consistent with public reports from OpenAI on their
internal DALL·E~3 attribution classifier, which indicate reduced performance
when distinguishing DALL·E~3 outputs from those of other AI models, with
approximately 5--10\% of non-DALL·E images flagged as DALL·E-generated.\footnote{
See \url{https://openai.com/index/understanding-the-source-of-what-we-see-and-hear-online/}.
}

One plausible explanation for the difficulty of these generators is that they
may share architectural components, training data, or stylistic characteristics
with DALL·E~3, leading to more similar feature representations. However, since
these models are closed-source, the precise causes of this overlap are difficult
to verify. In light of this, we focus our analysis on improving performance for
these harder cases.

\begin{table}[htbp]
\centering
\footnotesize
\caption{Average Precision (AP) and ROC AUC for attributing images to DALL·E~3
versus each comparison source, evaluated with and without wild data.
``SD'' denotes Stable Diffusion. ``cons.~opt.'' refers to constrained optimization
and ``pseudo'' to pseudo-labeling. The final column reports the average
performance over the challenging generators Midjourney, Firefly, and
Stable Diffusion~XL.}
\label{tab:main_results_short}
\begin{tabular}{lccccccc}
\toprule
\textbf{Metric} & \textbf{Wukong} & \textbf{SD v1.4} & \textbf{SD v1.5} & \textbf{Midjourney} & \textbf{Firefly} & \textbf{SD XL} & \textbf{Avg (hard)} \\
\midrule
\textit{AP (w/o wild)}  & 0.9959 & 0.9848 & 0.9974 & 0.9198 & 0.9284 & 0.8604 & 0.9029 \\
\textit{AP (pseudo)}  & \textbf{0.9967} & \textbf{0.9882} & \textbf{0.9979} & 0.9328 & 0.9436 & 0.8927 & 0.9230 \\
\textit{AP (cons. opt.)}   & 0.9936 & 0.9866 & 0.9957 & \textbf{0.9346} & \textbf{0.9472} & \textbf{0.9015} & \textbf{0.9278} \\

\midrule
\textit{AUC (w/o wild)} & 0.9957 & 0.9856 & 0.9974 & 0.9252 & 0.9259 & 0.8619 & 0.9043 \\
\textit{AUC (pseudo)}  & \textbf{0.9966} & \textbf{0.9882} & \textbf{0.9979} & 0.9340 & 0.9378 & 0.8930 & 0.9216\\
\textit{AUC (cons. opt.)}  & 0.9932 & 0.9866 & 0.9958 & \textbf{0.9361} & \textbf{0.9423} & \textbf{0.9033} & \textbf{0.9272} \\

\bottomrule
\end{tabular}
\end{table}

As shown in Table~\ref{tab:main_results_short}, incorporating wild data via
constrained fine-tuning leads to consistent improvements on the challenging
generators Midjourney, Firefly, and Stable Diffusion~XL. In particular, the
average AP across these generators increases from 0.9029 to 0.9278, and the
average ROC AUC increases from 0.9043 to 0.9272. At the same time, performance on
in-distribution generators remains high, with only minor fluctuations, indicating
that the introduction of wild data does not meaningfully degrade attribution
accuracy on known sources.

These results demonstrate the effectiveness of constrained fine-tuning for
leveraging unlabeled wild data to improve generalization to unseen and difficult
generators while preserving strong performance on labeled in-distribution data.

We attribute the observed gains to two complementary effects. First, the wild
data may contain samples from generators that overlap with the hard test cases,
providing additional exposure that improves discrimination. Second, exposure to
a broader and more diverse set of images helps the classifier learn a more robust
decision boundary separating DALL·E~3 outputs from non-target content, leading
to improved overall attribution performance.

\subsection{Pseudo-Labeling}

Our framework for incorporating wild data supports multiple strategies. In
addition to constrained optimization, we consider a pseudo-labeling baseline.
Under pseudo-labeling, the classifier iteratively assigns labels to wild samples
for which it produces high-confidence predictions (using a confidence threshold
of 90\%), and retrains on these pseudo-labeled examples. Training proceeds until
no additional confident pseudo-labels can be obtained. All other experimental
settings are identical to those used in the main experiment.

As shown in Table~\ref{tab:main_results_short} and in the full results reported
in the Appendix, both pseudo-labeling and constrained optimization yield strong
overall performance. Pseudo-labeling achieves slightly higher performance on
in-distribution generators, while constrained optimization consistently attains
higher AP and AUC on the challenging generators Midjourney, Firefly, and Stable
Diffusion~XL. These results suggest that pseudo-labeling primarily reinforces
patterns present in the labeled data, whereas constrained optimization is more
effective at shifting the decision boundary toward harder, previously unseen
cases.

\subsection{Comparison with Other Methods and Baselines}

To the best of our knowledge, this work is the first academic study to explicitly
address target generator attribution in the presence of unknown and unseen
generators, particularly for closed-source commercial models such as
DALL·E~3. Existing academic literature on AI-generated images has largely focused
on the binary real-versus-fake detection setting, which is fundamentally
different from the targeted attribution problem studied here and therefore not
directly comparable.

While industry systems for target generator attribution likely exist, their
designs and evaluation protocols remain proprietary and unpublished. Moreover,
prior academic work does not explore the use of unlabeled wild data for improving
targeted attribution robustness under open-world conditions. As such, there are
no directly comparable baselines that jointly address target generator
attribution, unknown generators, and wild-data utilization. We view this work as
a step toward filling this gap and hope it will help motivate future benchmarks
and methods for unknown-aware attribution.

\subsection{Ablation Studies}

We conduct a series of ablation studies to examine the robustness and behavior of
our approach under different training configurations. Across all settings, we
consistently observe that incorporating unlabeled wild data via constrained
optimization improves attribution performance on challenging generators while
preserving strong in-distribution (ID) performance.

\paragraph{Impact of Wild Data Size.}
We first study the effect of the wild dataset size while keeping the labeled ID
dataset fixed (DALL·E~3: 200 samples; Wukong, Stable Diffusion~v1.4, and v1.5:
67 samples each). Increasing the number of wild samples per generator from a
small to a suggested larger scale yields consistent, though moderate,
performance improvements, particularly for challenging generators such as
Midjourney, Firefly, and Stable Diffusion~XL.

These observations indicate that larger wild datasets can further enhance
generalization, although the marginal gains diminish as the dataset grows. This
behavior suggests that the linear classifier built on CLIP representations
already captures strong discriminative signals in low-data regimes, highlighting
the effectiveness of pretrained features for attribution.

Notably, even when the wild dataset substantially exceeds the size of the
labeled ID set, training remains stable. This stability arises from normalizing
losses on a per-sample basis and explicitly constraining the ID loss during
optimization, which together ensure robustness to dataset imbalance.

\paragraph{Impact of Mislabeled Target-Generator Samples in Wild Data.}
To assess robustness to label noise, we vary the proportion of target-generator
images (DALL·E~3) relative to other sources in the wild mixture, including an
extreme setting in which target and non-target images appear in roughly equal
proportions. Across all settings, the proposed approach consistently preserves
ID performance while improving attribution accuracy on challenging generators.

Even when a substantial fraction of wild data originates from the target
generator, attribution performance on hard sources improves relative to training
without wild data. AUC trends closely mirror AP improvements. These results
demonstrate that the constrained objective effectively mitigates the impact of
label noise and allows the model to benefit from wild data despite imperfect
source composition.

\paragraph{Effect of Larger Labeled ID Sets.}
We next examine whether the benefit of incorporating wild data diminishes as more
labeled training data becomes available. We consider progressively larger ID
datasets, ranging from moderately expanded settings to substantially larger
ones with thousands of labeled samples per generator.

Across these regimes, fine-tuning with wild data continues to yield improvements,
particularly for challenging generators such as Midjourney and Stable
Diffusion~XL. This suggests that the proposed framework remains effective beyond
low-data scenarios and provides a scalable mechanism for leveraging unlabeled
data to improve generalization.

\paragraph{Varying Generator Diversity in the ID Dataset.}
To further assess robustness across training configurations, we vary the
composition of the labeled ID dataset by including either real images (e.g.,
from ImageNet) or challenging generators such as Midjourney directly in
supervised training. While one might expect that exposure to such diverse or
difficult sources would reduce the marginal utility of wild data, we find that
fine-tuning with wild data continues to improve generalization performance.

These results suggest that wild data provides complementary coverage beyond the
labeled set, likely due to its broader and more heterogeneous distribution.

\paragraph{Effect of the ID Loss Constraint Threshold.}
Finally, we study the effect of relaxing the constraint on the labeled ID loss.
While our primary experiments constrain the ID loss to be at most twice its
baseline value, allowing larger deviations leads to stronger gains on
challenging generators at the cost of slightly reduced performance on some
other sources.

This trade-off highlights the flexibility of our framework: by adjusting the
constraint threshold, practitioners can explicitly balance improved
generalization to unknown generators against stricter preservation of
in-distribution performance, depending on deployment requirements.

\subsection{Discussion}
\label{sec:discussion}

Our experiments show that constrained fine-tuning with unlabeled wild data
consistently improves attribution robustness to unseen and challenging image
generators, even in low-label regimes. The proposed approach remains effective
across a broad range of training configurations, scales favorably with the
availability of wild data, and compares favorably with alternative strategies
such as pseudo-labeling. These properties make it well suited for realistic
deployment settings in which generative models evolve rapidly and labeled data
is expensive to obtain.

Several observations emerge from our empirical analysis:
\begin{itemize}
    \item \textbf{Stability with large wild datasets.}
    By normalizing losses on a per-sample basis and explicitly constraining the
    in-distribution (ID) loss, the optimization procedure remains stable even
    when the wild dataset substantially exceeds the size of the labeled ID set.

    \item \textbf{Robustness to label noise in wild data.}
    Treating wild samples as non-target inevitably introduces label noise, since
    some wild images may originate from the target generator. Nevertheless, the
    explicit constraint on ID loss mitigates overfitting to such noise, allowing
    the model to preserve accuracy on labeled data while benefiting from
    exposure to diverse wild samples.

    \item \textbf{Controllable trade-off via constraint tuning.}
    Relaxing the constraint on the labeled ID loss (e.g., from $2\times$ to
    $3\times$ the baseline loss) enables a principled trade-off between
    preserving in-distribution performance and improving generalization to
    unknown generators. This flexibility allows the method to be adapted to
    deployment-specific priorities.

    \item \textbf{Comparison with pseudo-labeling.}
    Pseudo-labeling provides a natural alternative within our semi-supervised
    framework, in which confident predictions on wild data are iteratively
    incorporated as labels. While pseudo-labeling slightly improves performance
    on in-distribution generators, constrained optimization consistently yields
    larger gains on the most challenging unseen generators, suggesting that it
    more effectively pushes the decision boundary toward harder cases.
\end{itemize}

\paragraph{Limitations.}
The effectiveness of the proposed approach depends on the diversity of the wild
data. When the wild dataset is highly biased or lacks coverage of challenging
sources, the resulting generalization gains may be limited. In addition, our
experiments rely on CLIP ViT-L/14 as the feature backbone; exploring alternative representation models may further improve attribution performance. Finally, we do not address robustness to adversarial manipulations or intentional obfuscation of generator-specific signatures, which we leave as an important direction for future work.

\section{Conclusion}

We investigate the problem of improving target generator attribution by leveraging unlabeled wild data. Although CLIP-based representations combined
with a linear classifier achieve strong performance on many sources, accurately
distinguishing a target generator from certain challenging generators, such as
Midjourney, Firefly, and Stable Diffusion~XL, remains difficult. To address this
challenge, we propose a constrained optimization framework that incorporates
unlabeled wild images while explicitly preserving performance on labeled
in-distribution data.

Extensive experiments demonstrate that the proposed approach consistently
improves generalization to unseen and difficult generators without sacrificing
accuracy on known sources. These gains are robust across varying wild data
sizes, labeled data configurations, and alternative semi-supervised strategies
such as pseudo-labeling. By normalizing losses on a per-sample basis and
explicitly constraining in-distribution performance, the method remains stable
even when the volume of wild data substantially exceeds that of labeled data.

More broadly, our framework enables the effective use of unlabeled data without requiring manual relabeling and supports incremental fine-tuning as new generative models emerge. These properties make it well suited for realistic, open-world image attribution scenarios in which the generator landscape evolves rapidly.

\clearpage

\bibliography{reference}
\bibliographystyle{unsrt}

\end{document}